\pdfoutput=1
\documentclass[sigconf]{acmart}
\usepackage{booktabs} % For formal tables
\usepackage{microtype}
\usepackage{overpic}

\setcopyright{none}
\acmConference[ICVGIP'24]{15th Indian Conference on Computer Vision, Graphics and Image Processing}{December 2024}{Bangalore, India}
\acmYear{2024}
\begin{document}
%\title{Affordable Non-Metameric Reconstruction of Illuminant Spectra Using CD-ROM }
%\title{Reconstructing Spectral Power Distribution of an Illuminant Using A CD-ROM}

\title{Practical and Accurate Reconstruction of an Illuminant's Spectral Power Distribution for Inverse Rendering Pipelines}

% \titlenote{Produces the permission block, and
  % copyright information}

\author{Parisha Joshi}
\affiliation{%
  \institution{Clemson University}
  % \streetaddress{Anonymous}
  \city{Clemson}
  \state{SC}
  \country{USA}
  \postcode{29634}
}

\author{Daljit Singh J. Dhillon}
\affiliation{%
  \institution{Clemson University}
  % \streetaddress{Anonymous}
  \city{Clemson}
  \state{SC}
  \country{USA}
  \postcode{29634}
}

% \renewcommand{\shortauthors}{Anonymous}

\begin{abstract}
%Alongside material properties, shape/geometry, and camera parameters, knowing illumination spectra adds to the scene's realism.
Inverse rendering pipelines are gaining prominence in realizing photo-realistic reconstruction of real-world objects for emulating them in virtual reality scenes. %for computer graphics based content creation. %that are hypothetically placed in virtual scene-graphs. 
%while modeling and simulating them in virtual reality systems.% for rich immersive user experiences.
Apart from material reflectances, spectral rendering and in-scene illuminants' spectral power distributions (SPDs) play important roles in producing photo-realistic images. We present a simple, low-cost technique to capture and reconstruct the SPD of uniform illuminants. Instead of requiring a costly spectrometer for such measurements, our method uses a diffractive compact disk (CD-ROM) and a machine learning approach for accurate estimation.  We show our method to work well with spotlights under simulations and few real-world examples. Presented results clearly demonstrate the reliability of our approach through quantitative and qualitative evaluations, especially in spectral rendering of iridescent materials. %In future, we will adapt this method for real-world acquisition. 
\end{abstract}

\keywords{Inverse rendering, spectral illumination, diffraction, SPDs, RNNs}

\begin{teaserfigure}
%    \centering
\setlength{\tabcolsep}{1pt}
\begin{tabular}{ccccc}
    \includegraphics[width=0.172\textwidth]{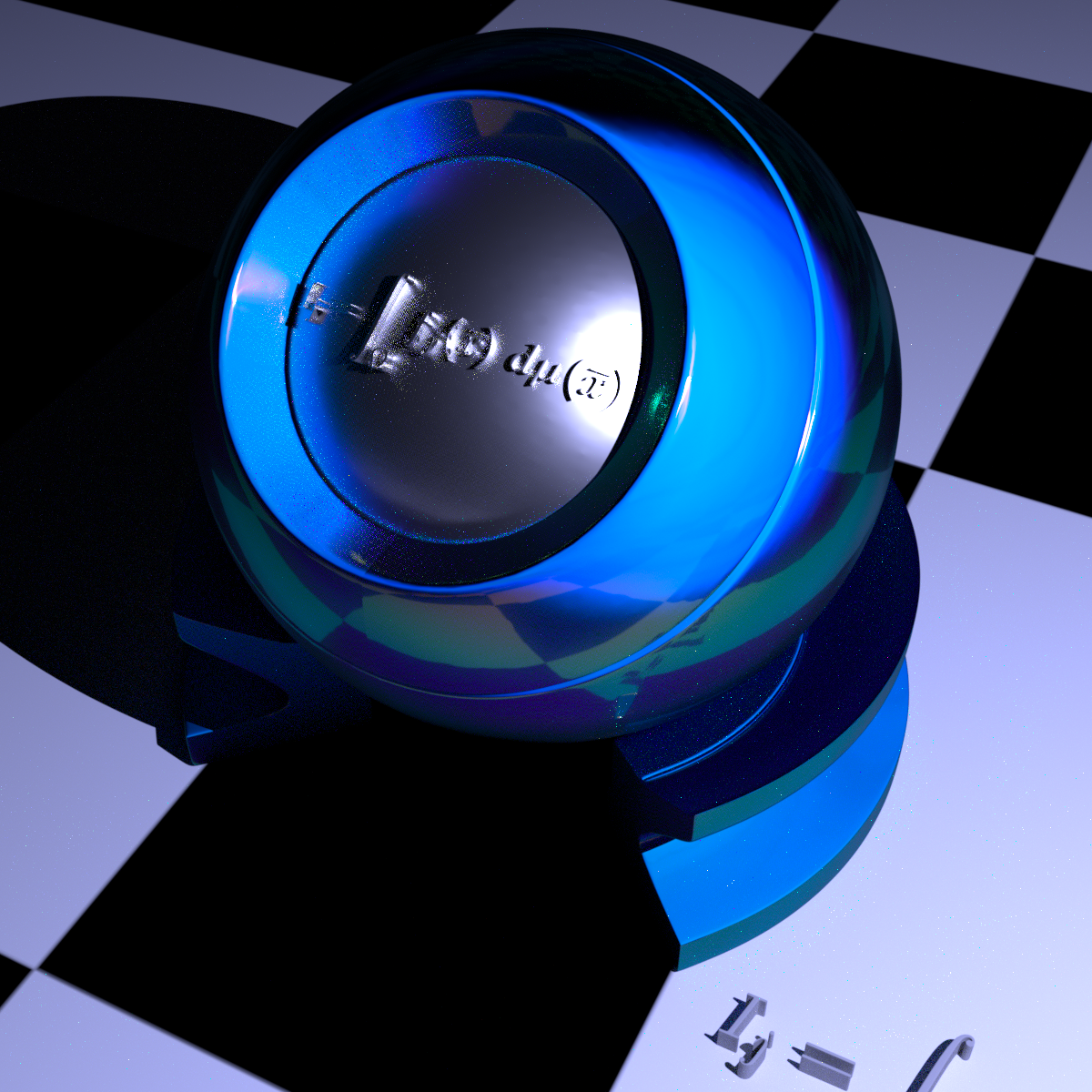}&
    \includegraphics[width=0.172\textwidth]{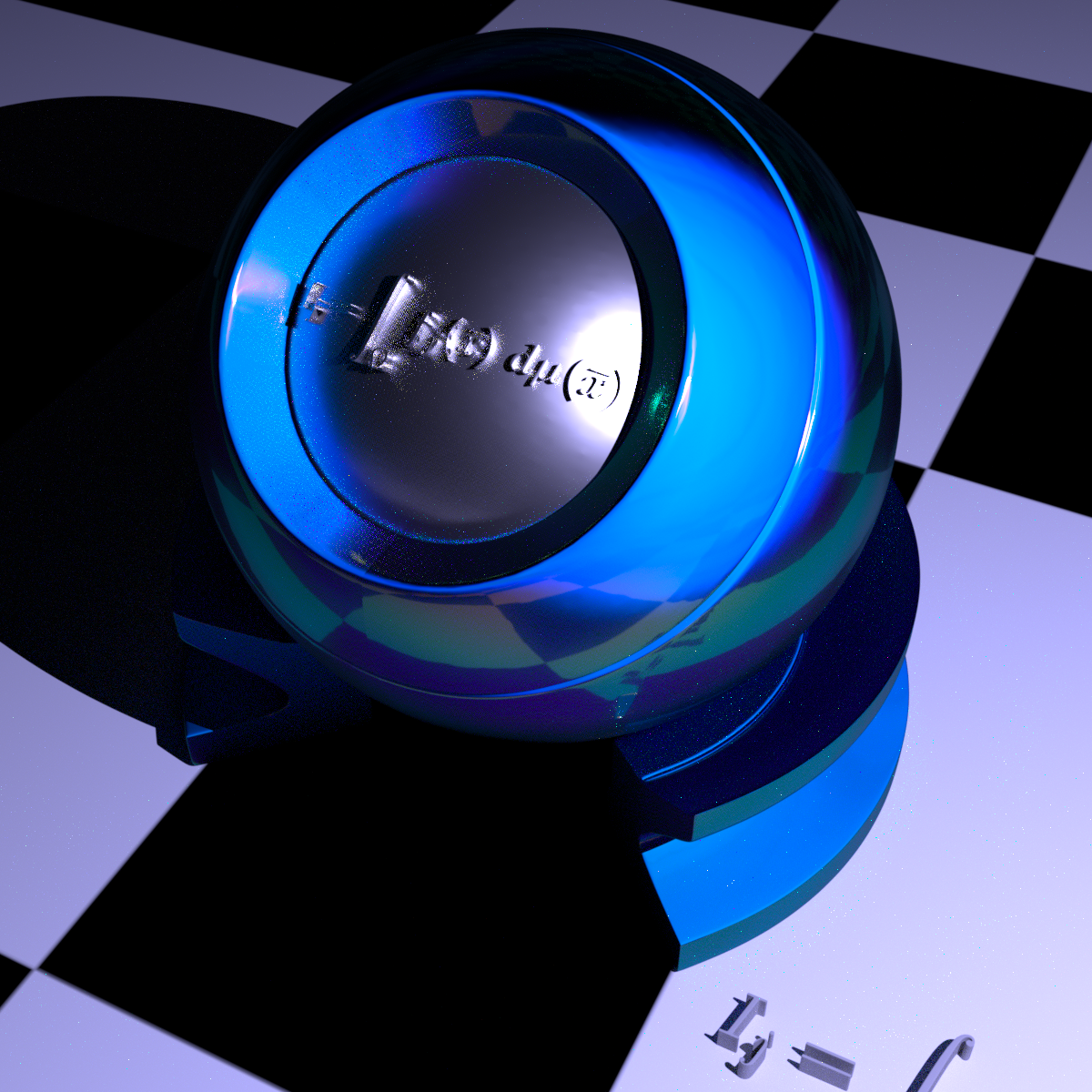}&
    \includegraphics[width=0.172\textwidth]{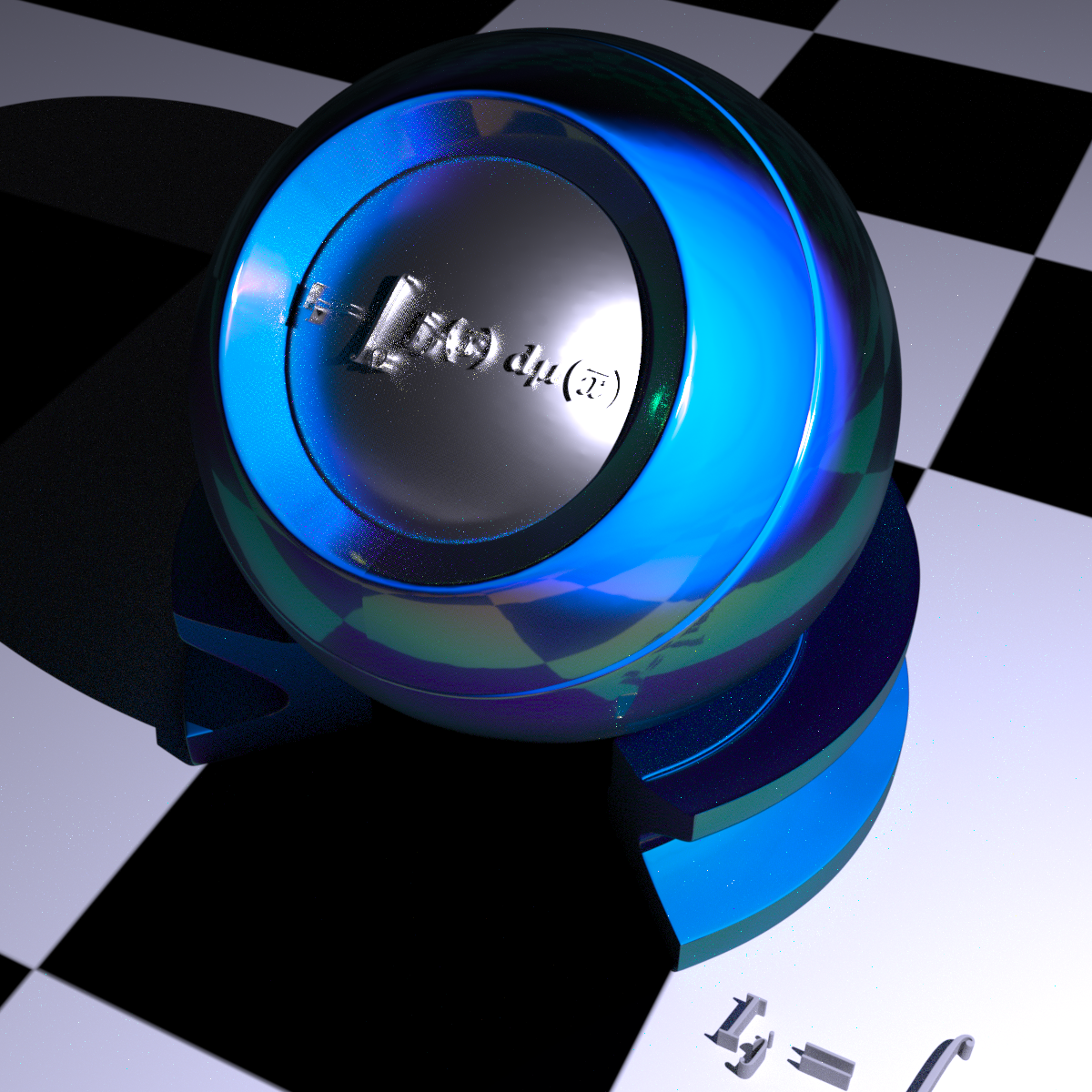}&
      \includegraphics[height=0.172\textwidth]{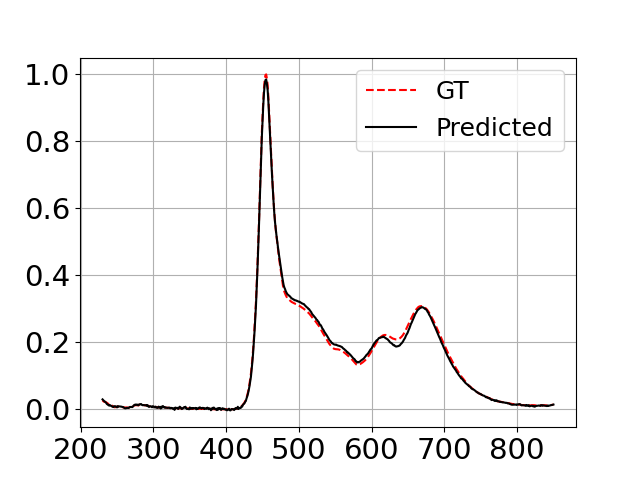}&
     \includegraphics[height=0.172\textwidth]{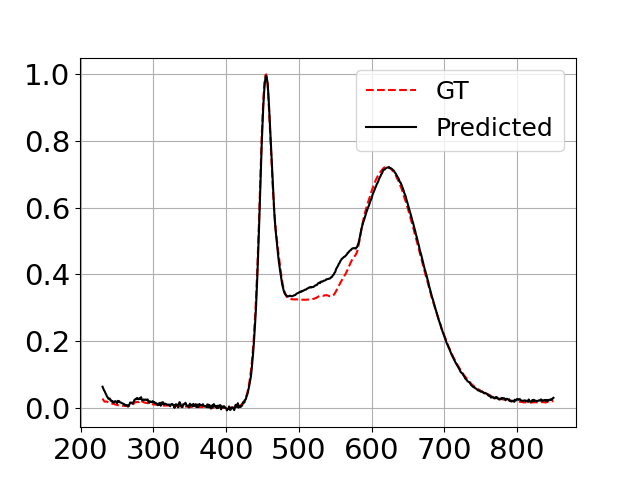}
\\
    \includegraphics[width=0.172\textwidth]{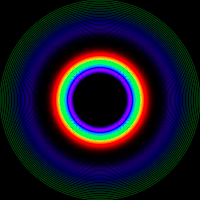}&
    \includegraphics[width=0.172\textwidth]{11_PD_CD_a_1500_h_15.png}&
    \includegraphics[height=0.172\textwidth]{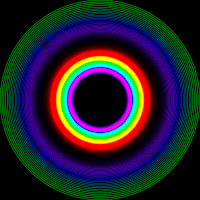}&
   \includegraphics[height=0.172\textwidth]{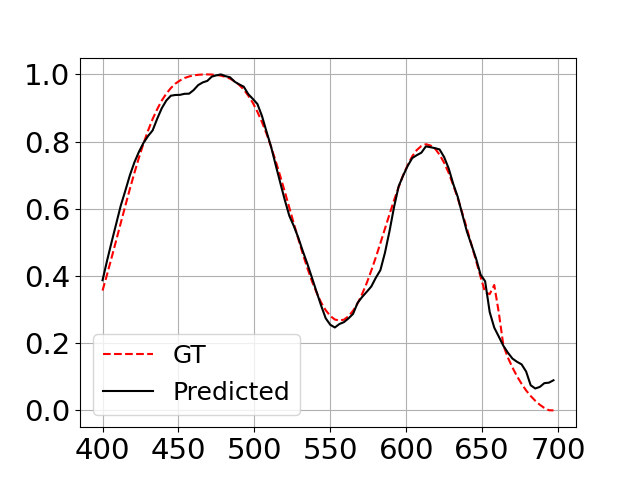}&
    \includegraphics[height=0.172\textwidth]{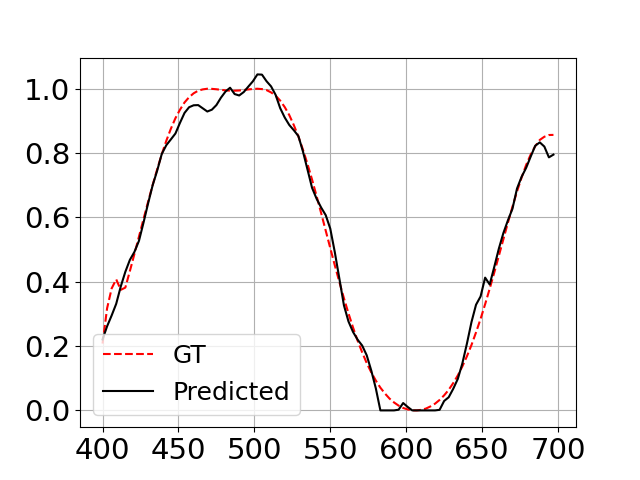}\\
    (a) Renderings: Ours\hfill{} & Ground truth & Reference RGB  & \multicolumn{2}{c}{ (b)  Few predicted spectra in comparison with their \textit{ground truths}}
    \\ 
%    &\textbf{(a)}&&\multicolumn{2}{c|}{\textbf{(b)}} 
\end{tabular}
\Description{(a) We use a measured iridescent material BRDF \cite{dupuy&jacob} and a simulated CD grating's BRDF to validate our reconstructed spectra against the ground truth under rendering tasks. (b) Shows predicted and ground truth SPDs for four illuminants. X-axes represent the wavelengths in nanometers and Y-axes show relative intensities. The bottom two are synthetic and the top two are real world illuminants. Ground truth for them was acquired using a high-end spectrometer (Hopoocolor OHSP350UV 230 to 850nm). The top left spectral profile was used to generate iridescent and CD rendering in figure (a). The reference RGB renderings exhibit visible mismatches in color tones as well as relative intensities. In comparison, our reconstructed spectra produce renderings that are qualitatively similar to corresponding ground truths.}
\caption{ (a) We use a measured iridescent material BRDF \cite{dupuy&jacob} and a simulated CD grating's BRDF to validate our reconstructed spectra against the ground truth under rendering tasks. (b) Shows predicted and ground truth SPDs for four illuminants. X-axes represent the wavelengths in nanometers and Y-axes show relative intensities. The bottom two are synthetic and the top two are real world illuminants. Ground truth for them was acquired using a high-end spectrometer (Hopoocolor OHSP350UV $230$\textendash $850$nm). The top left spectral profile was used to generate iridescent and CD rendering in figure (a). The reference RGB renderings exhibit visible mismatches in color tones as well as relative intensities. In comparison, our reconstructed spectra produce renderings that are qualitatively similar to corresponding ground truths.}
    % Caption and description
    % \caption{Grid of images with 3 rows and 4 columns.}
    % \Description{A grid containing twelve images arranged in three rows and four columns. Each image is represented as a part of the grid.}
\end{teaserfigure}

\maketitle

\section{Background}
Spectral rendering pipelines that produce high-quality, photo-realistic images require material properties and illumination sources to be represented with accurate spectral characterization. Recent advances in inverse rendering have mainly focused on multi-spectral acquisition of material appearance data~\cite{legendre2016,legendre2018} for data-driven relighting, indirect environmental light maps~\cite{guarnera2022} and parametric, spectral reflectance (BRDFs)~\cite{dupuy2018} or scattering (BSDFs)~\cite{gitlina2020} functions, in this context. %For recreating light sources from real-world examples, VFX artists and animators generally rely on expensive, handheld spectrometers to    
In this paper, instead, we focus on accurate reconstruction of the uniform spectral power distribution (SPD) of common real-world illuminants that a content-creators may wish to replicate in their virtual setups. These SPDs are critical for accurate reconstruction of iridescence and structural coloration effects such as those on insect and reptile bodies~\cite{dhillon2014}, scratched or glinty surfaces~\cite{werner2017scratch,yan2018rendering}, Bragg mirrors~\cite{fourneau2024}, layered materials with specular sheens~\cite{belcour2017practical}. While spectrometers exist to measure illuminant SPDs accurately, they are expensive and not easily available to most artists. We thus devise a simple, effective and affordable method that can be adapted for any camera with one known light source and a set of known transmissive color filters. 
%\vspace{-2mm}
\begin{figure}[t]
    \centering
     \begin{Overpic}{%
     \includegraphics[width=0.65\linewidth]{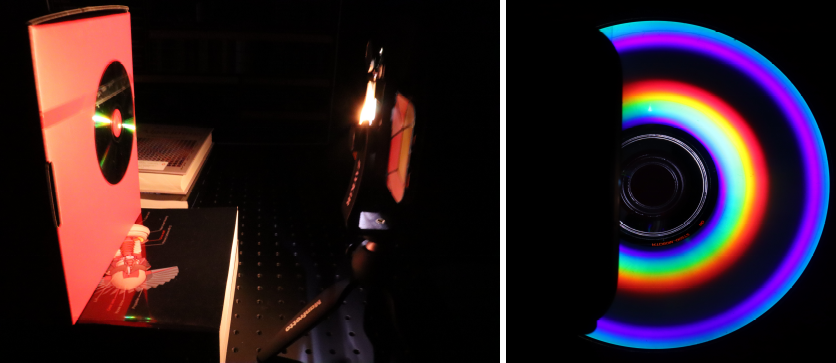}
     \includegraphics[width=0.34\linewidth]{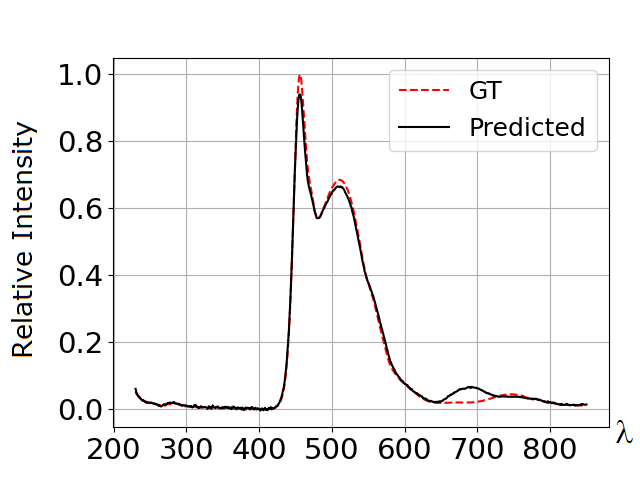}
    }%
    %\put (34,25){\textbf{{\color{white}(a)}}}%
    %\put (59,25){\textbf{{\color{white}(b)}}}%
    %\put (64,25){\textbf{\colorbox{white}{\hspace{1mm}(c)}}}%
    \put (-0.2,25){\textbf{\colorbox{white}{\color{black}(a)}}}%
    \put (38.6,25){\textbf{\colorbox{white}{\color{black}(b)}}}%
    %\put (94,25){\textbf{\colorbox{white}{(c)}}}%
    \put (66,25){\textbf{\colorbox{white}{\hspace{1mm}(c)}}}%
    
    %\put (2,25){\textbf{{\color{yellow}(a)\hspace{2.9cm}(b)\hspace{1.7cm}}}\colorbox{blue!30}{(c)}}%
  \end{Overpic}%
     % \includegraphics[width=0.3\linewidth]{icvgip2021-latex-template/Real_world_CD/diffraction_00_2024_09_04_IMG_0073_half.JPG}
     %\includegraphics[width=0.3\linewidth]{icvgip2021-latex-template/Real_world_CD/73_illuminant_real_world.png}\\
     %\textbf{(a)} \hspace{0.13\textwidth}\textbf{(b)}\hspace{0.13\textwidth}\textbf{(c)}
    \vspace{-2mm}
    \Description{(a) Real World CD Capture setup with Canon Rebel T8i (b)With few adjustments of CD or illumination the rings were captured with the SPD shown in (c) measured with Hoppoocolor OHSP350 .}
    \caption{(a) Real World CD Capture setup with Canon Rebel T8i (b)With few adjustments of CD or illumination the rings were captured with the SPD shown in (c) measured with Hoppoocolor OHSP350 .
    }
    \label{fig:enter-label}
    \vspace{-0.5cm}
\end{figure}

\section{Proposed Method}
Spectrometers~\cite{zeiss} and hyper-spectral imaging methods \cite{minkim} commonly rely on a diffractive optical element (filter) to profile the spectral distribution of the incident radiance/s. This inspired us to use a simple, diffractive compact disk (CD-ROM) in devising an image-based method for spectral profiling of light sources. CD-ROM are inexpensive, standardized, high-quality grating constructs that are also easily available. Thus, using them as the essential diffractive element has allowed us in devising a simple, reliable and cost-effective method for estimating the SPDs of light sources.      

\noindent\textbf{\textit{Imaging Setup:\hspace{2mm}}} We have experimented in a simulated imaging environment (PBR toolkit~\cite{pharr2023}) that supports spectral rendering. In the next section, we discuss real-world, practical adaption of our method along with a few results. We illuminate an unwritten CD-ROM with a spotlight that is placed fronto-parallelly to it and capture its appearance with a camera that is also placed fronto-parallelly to the CD. The optical axis of the camera passes through the CD's center. The unwritten CDs have fixed circular tracks as a single diffractive layer of known spacing between the gratings. We simulate the reflectance function for the CD gratings by implementing the analytic BRDF given by Toisoul et~ al.~\cite{toisoul2018acquiring}.  We set the grating gap to $a=0.5\mu m$ and its maximum height as $h0 = 0.15 \mu m$. Our synthetic camera has known color filter functions and its exposure settings are fixed to avoid color and brightness saturation. Using this setup, we developed a data-driven model for estimating the SPD of any unknown spotlight in exactly the same configuration.

\noindent\textbf{\textit{Machine Learning Step:\hspace{2mm}}}
We illuminate this CD with a set of synthetic SPDs to produce a set of images. The SPDs are generated to smooth variations, random noise as well as sharp spikes in different combinations to mimic a large variety of real world SPDs. Each SPD is put on a relative scale to have its peak value as $1$. With the data set corresponding of 5000 SPD we train a multilayer perceptron (MLP) network for learning the supervised regression process.
%We simulated a simple compact disk using pbrt renderer with measurement approach given by \cite{dupuy&jacob} for diffracted surfaces. The circular gratings are created by procedurally considering the distance between two gratings $a=0.5\mu m$ and height field $h0 = 0.15 \mu m$. For the illumination profiles, there were very limited number of spds available for real world illuminants. Hence, synthetic SPDs were generated with random noise mimicking real world SPDs. With the data of 5000 SPD profiles and 5000 compact disk images the machine learning model was trained. 
%We used a simple but effective multilayer perceptron to convert the CD image to latent space and then decompress to make a visual spectra. 
Each image is pre-processed to mask out the inner and outer circular regions around the CD that do not contribute to the learning process. With Adam optimizer and leaky RELU activation function, the model generally requires training for up to $100000$ epochs. We use a batch size of $64$. We train with $4000$ random SPD samples and validate against the remaining $1000$ SPD samples.

\noindent\textbf{\textit{Results:\hspace{2mm}}}
We performed quantitative evaluations using statistical measures such as MAE, RMSE, and the correlation factor between the ground truth SPD and the respective prediction. Average values across the validation set for all these measures are shown in Table~1. The last column clearly indicates that our MLP network predicts unknown SPDs with very high accuracy. We also, compared our method's performance under rendering. Figure 1 and Figure 3 show that our renderings are visually indistinguishable from the ground truth. Rendering PSNRs are also indicated in Figure~3.

%The results are shown in Table 1 for training with 4000 samples and validating with 1000 samples. The same process of measuring the lengths of each specular lobe manually would have been a rigorous task. 

% \begin{figure}
%     \centering
%      \includegraphics[width=0.3\linewidth]{icvgip2021-latex-template/Real_world_CD/IMG_0327.jpeg}
%      \includegraphics[width=0.3\linewidth]{icvgip2021-latex-template/Real_world_CD/diffraction_00_2024_09_04_IMG_0073_half.JPG}
%      \includegraphics[width=0.3\linewidth]{icvgip2021-latex-template/Real_world_CD/73_illuminant_real_world.png}\\
%      \textbf{(a)} \hspace{0.2\textwidth}\textbf{(b)}
%     \caption{Real World CD Capture with Canon Rebel T8i.}
%     \label{fig:enter-label}
% \end{figure}
% \begin{figure}[h]
%   \centering
%   \begin{subfigure}{0.45\linewidth}
%     \centering
%     \includegraphics[width=\linewidth]{icvgip2021-latex-template/Real_world_CD/IMG_0327.jpeg}
%     \caption{(a)}
%   \end{subfigure}
%   \hfill
%   \begin{subfigure}{0.45\linewidth}
%     \centering
%     \includegraphics[width=\linewidth]{icvgip2021-latex-template/Real_world_CD/IMG_0298.jpeg}
%     \caption{(b)}
%   \end{subfigure}
%   \caption{Real World CD Capture with Canon Rebel T8i.}
%   \label{fig:real}
% \end{figure}

\begin{figure}[t]
\Description{}
\setlength{\tabcolsep}{1pt}
% \begin{tabular}{@{\hspace{}}c@{\hspace{0.001\textwidth}}c@{\hspace{}}c@{\hspace{}}}
\begin{tabular}{@{\hspace{0.0mm}}c@{\hspace{0.2mm}}c@{\hfill{}}c@{\hspace{0.0mm}}} 
    \includegraphics[width=0.143\textwidth]{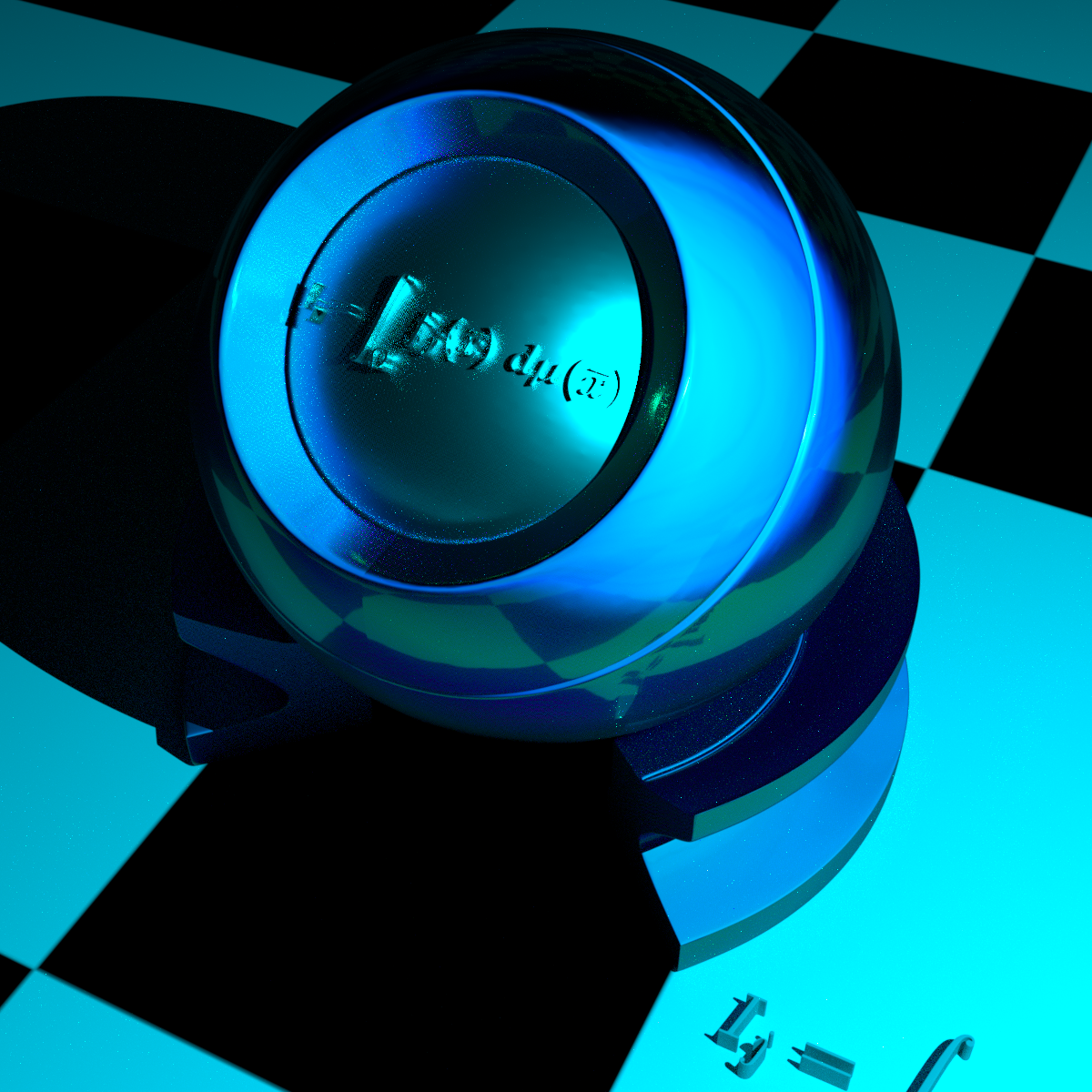}&
   \includegraphics[width=0.143\textwidth]{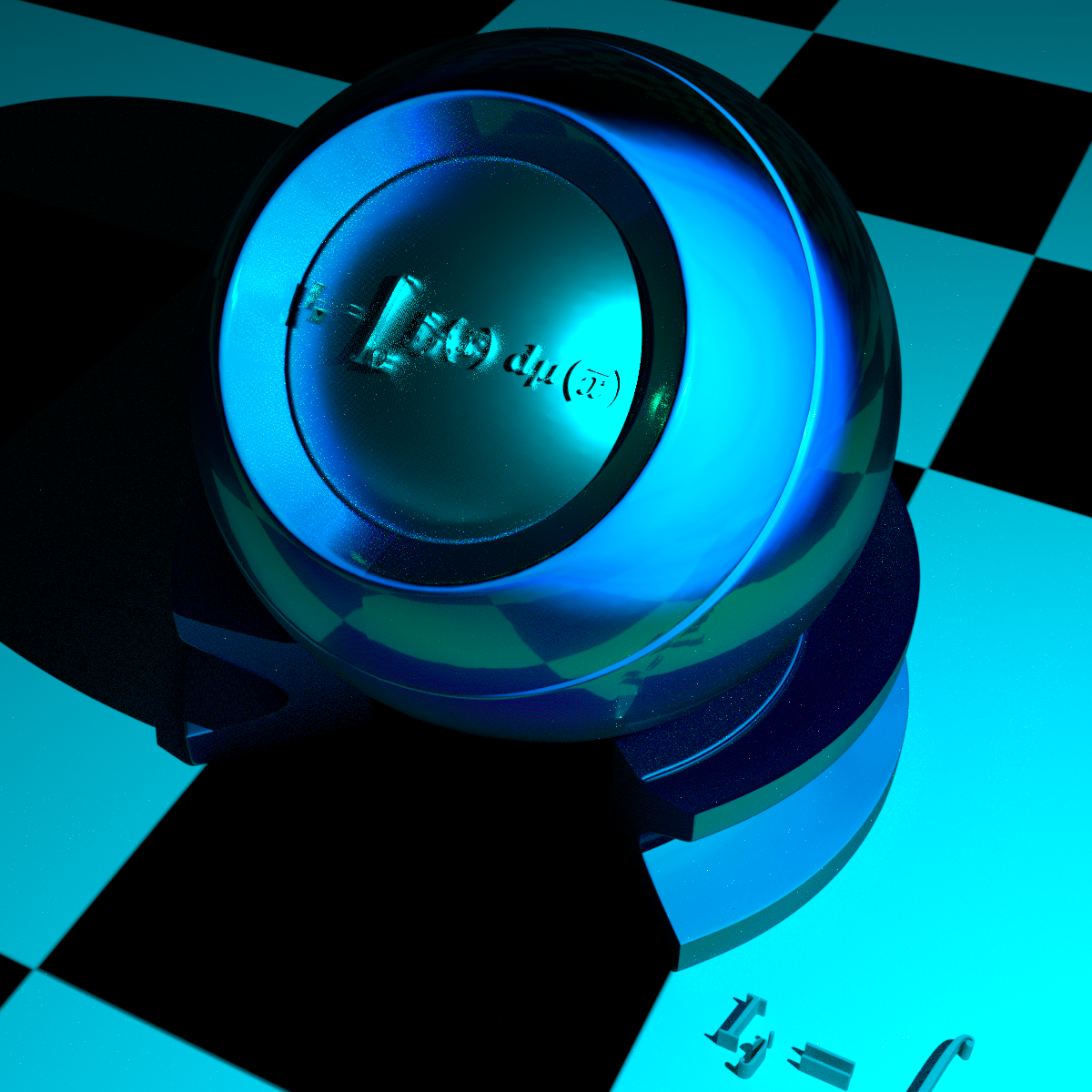} &
    \includegraphics[width=0.19\textwidth]{35_illuminant.png}\\
    \includegraphics[width=0.143\textwidth]{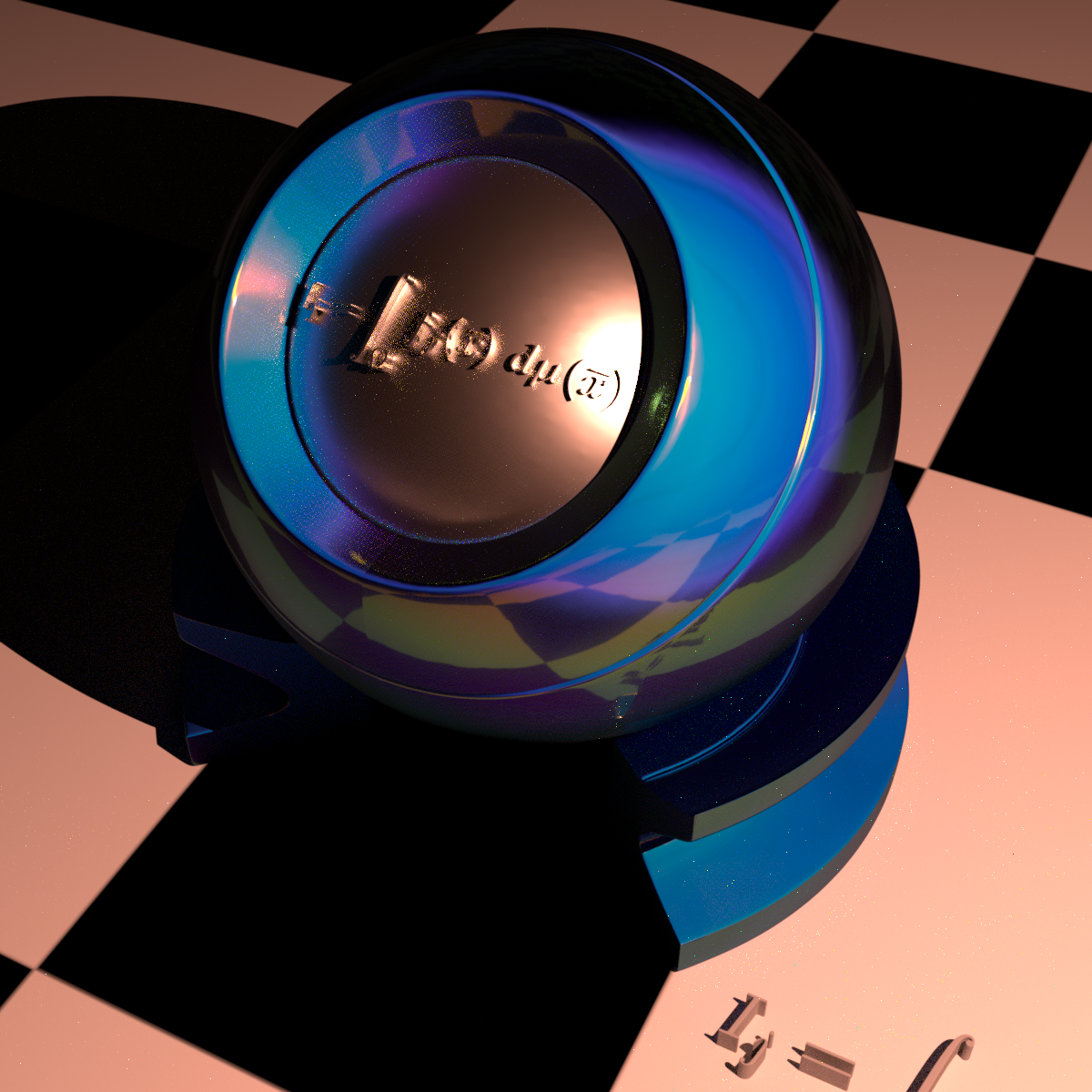} &
    \includegraphics[width=0.143\textwidth]{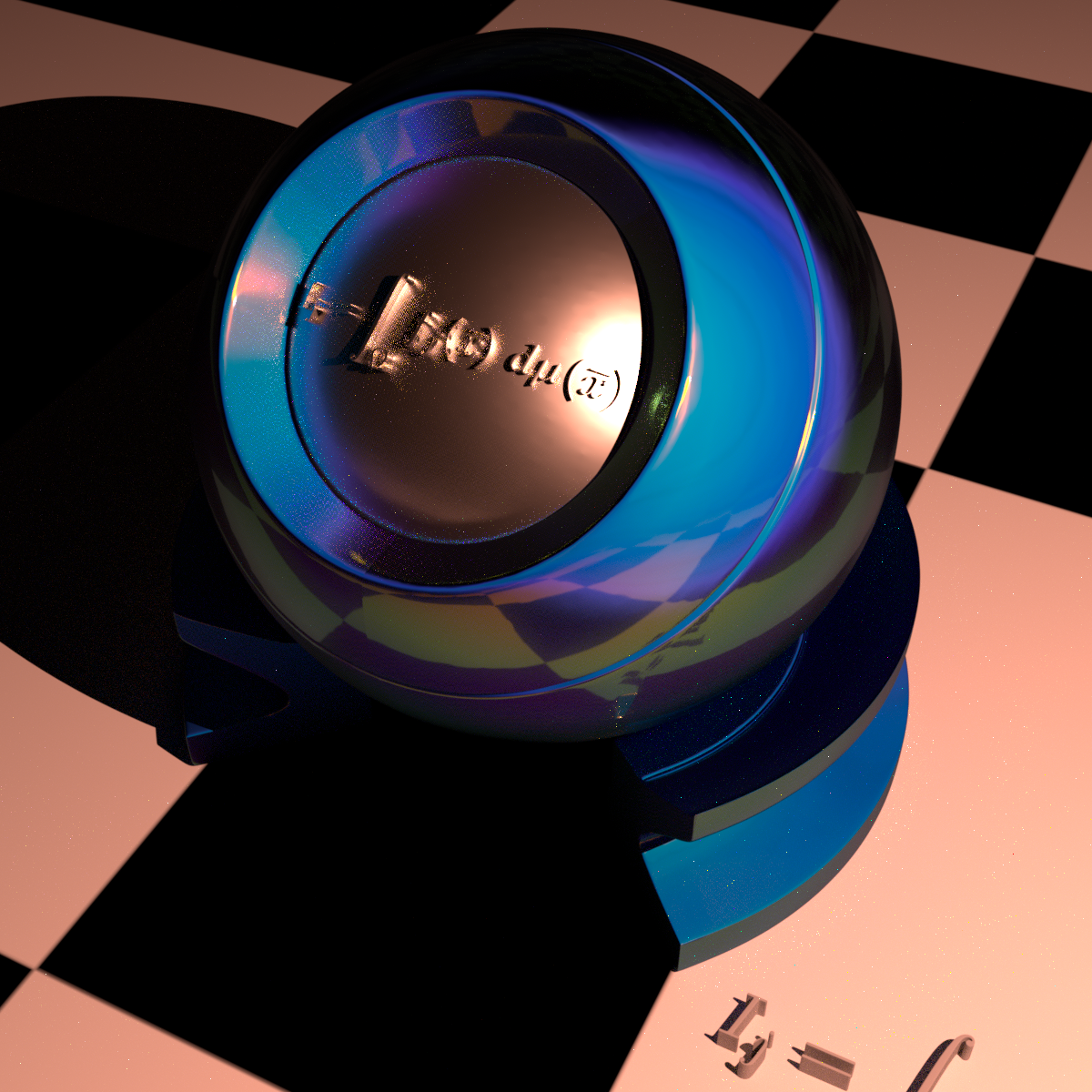} &    
    \includegraphics[width=0.19\textwidth]{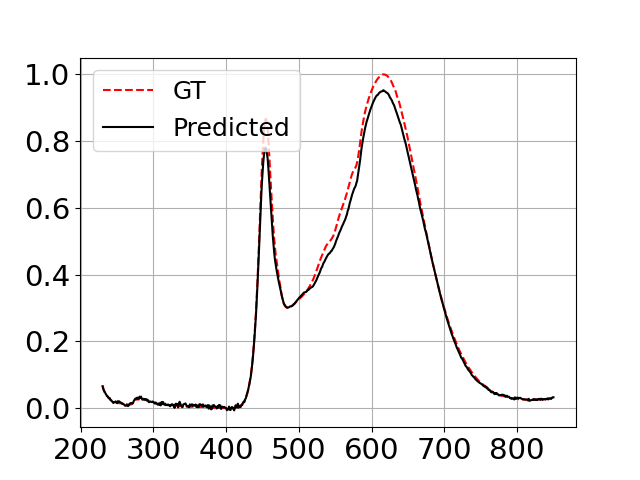}\\
    Ground Truth & Ours & Spectrum\\
\end{tabular}
% \Decription{This figure shows two more examples in SPD reconstruction and the impact on rendering accuracy.}
\vspace{-3mm}
\caption{ Two more examples in SPD reconstruction and the impact on rendering accuracy. Rendering PSNR for top = 46.596 dB and for the bottom = 55.352 dB}
\end{figure}

\begin{table}[t]
\vspace{-0.1cm}
\caption{Results of the MLP training step}
  \label{tab:freq}
\vspace{-2mm}
\vspace{-0.2cm}
  \begin{tabular}{|c|c|c|}
    \hline
    Metric& Training& Validation\\
    (Average) &(4000 sample SPDs) &(1000 unknown SPDs)\\
    \hline
    % MSE & 5.9034e-03 & 0.013518 \\
    MAE & 0.0466 & 0.06771 \\
    RMSE & 0.007099 & 0.0105 \\
    % PSNR & 23.49855 & 20.3911 \\
    % Canberra & 16.81028 & 19.9367 \\ 
    % SAM & 0.15493 &  0.2323 \\
    Correlation  & 0.936503 & 0.86411 \\
    % Euclidean & 0.70994& 1.0506 \\
  \hline
\end{tabular}
\vspace{-.5cm}
\end{table}

\section{Discussion and Future Work}
The training and validation results conclude that the spotlight illumination spectra can be reconstructed with minimal error using diffraction CD, under simulations. We also experimented to adapt our method for real-world illuminants. Figure 2 shows our imaging setup, one example case and the resulting SPD. For training with the real-world SPDs, we used the spectrometer to obtain the ground truth. We find out method to produce promising initial results that match up to our findings under PBRT simulations. In the future, we want to thoroughly validate our method against a large set of real-worlds images. Also, we would want to extend our method to work with any camera that is color-calibrated using a single MLP. 
%Also we conjeture that our model can be generalized to address illuminana at varying distances.
Lastly, we conjecture that our method can be adapted to work with general placements and non-uniform illuminants. Such improvements have the potential to support spectral environment maps for photo-realistic rendering of nuanced structural colorations.\\[-5pt]

\noindent \textbf{ACKNOWLEDGEMENTS.\hspace{1mm}} This work is fully supported by NSF Grant \#2007974.

\appendix


% The appendix gets added after the references.




\begin{thebibliography}{15}

%%% the Web address.

\ifx \showCODEN    \undefined \def \showCODEN     #1{\unskip}     \fi
\ifx \showDOI      \undefined \def \showDOI       #1{#1}\fi
\ifx \showISBNx    \undefined \def \showISBNx     #1{\unskip}     \fi
\ifx \showISBNxiii \undefined \def \showISBNxiii  #1{\unskip}     \fi
\ifx \showISSN     \undefined \def \showISSN      #1{\unskip}     \fi
\ifx \showLCCN     \undefined \def \showLCCN      #1{\unskip}     \fi
\ifx \shownote     \undefined \def \shownote      #1{#1}          \fi
\ifx \showarticletitle \undefined \def \showarticletitle #1{#1}   \fi
\ifx \showURL      \undefined \def \showURL       {\relax}        \fi
% invisible to TeX
\providecommand\bibfield[2]{#2}
\providecommand\bibinfo[2]{#2}
\providecommand\natexlab[1]{#1}
\providecommand\showeprint[2][]{arXiv:#2}

\bibitem[\protect\citeauthoryear{Belcour and Barla}{Belcour and Barla}{2017}]%
        {belcour2017practical}
\bibfield{author}{\bibinfo{person}{Laurent Belcour} {and} \bibinfo{person}{Pascal Barla}.} \bibinfo{year}{2017}\natexlab{}.
\newblock \showarticletitle{A practical extension to microfacet theory for the modeling of varying iridescence}.
\newblock \bibinfo{journal}{\emph{ACM Transactions on Graphics (TOG)}} \bibinfo{volume}{36}, \bibinfo{number}{4} (\bibinfo{year}{2017}), \bibinfo{pages}{1--14}.
\newblock


\bibitem[\protect\citeauthoryear{Dhillon, Teyssier, Single, Gaponenko, Milinkovitch, and Zwicker}{Dhillon et~al\mbox{.}}{2014}]%
        {dhillon2014}
\bibfield{author}{\bibinfo{person}{D.~S. Dhillon}, \bibinfo{person}{J. Teyssier}, \bibinfo{person}{M. Single}, \bibinfo{person}{I. Gaponenko}, \bibinfo{person}{M.~C. Milinkovitch}, {and} \bibinfo{person}{M. Zwicker}.} \bibinfo{year}{2014}\natexlab{}.
\newblock \showarticletitle{Interactive Diffraction from Biological Nanostructures}.
\newblock \bibinfo{journal}{\emph{Computer Graphics Forum}} \bibinfo{volume}{33}, \bibinfo{number}{8} (\bibinfo{year}{2014}), \bibinfo{pages}{177--188}.
\newblock
\urldef\tempurl%
\url{https://doi.org/10.1111/cgf.12425}
\showDOI{\tempurl}


\bibitem[\protect\citeauthoryear{Dupuy and Jakob}{Dupuy and Jakob}{2018a}]%
        {dupuy&jacob}
\bibfield{author}{\bibinfo{person}{Jonathan Dupuy} {and} \bibinfo{person}{Wenzel Jakob}.} \bibinfo{year}{2018}\natexlab{a}.
\newblock \showarticletitle{An adaptive parameterization for efficient material acquisition and rendering}.
\newblock \bibinfo{journal}{\emph{ACM Trans. Graph.}} \bibinfo{volume}{37}, \bibinfo{number}{6}, Article \bibinfo{articleno}{274} (\bibinfo{date}{dec} \bibinfo{year}{2018}), \bibinfo{numpages}{14}~pages.
\newblock
\showISSN{0730-0301}
\urldef\tempurl%
\url{https://doi.org/10.1145/3272127.3275059}
\showDOI{\tempurl}


\bibitem[\protect\citeauthoryear{Dupuy and Jakob}{Dupuy and Jakob}{2018b}]%
        {dupuy2018}
\bibfield{author}{\bibinfo{person}{Jonathan Dupuy} {and} \bibinfo{person}{Wenzel Jakob}.} \bibinfo{year}{2018}\natexlab{b}.
\newblock \showarticletitle{An adaptive parameterization for efficient material acquisition and rendering}.
\newblock \bibinfo{journal}{\emph{ACM Transactions on graphics (TOG)}} \bibinfo{volume}{37}, \bibinfo{number}{6} (\bibinfo{year}{2018}), \bibinfo{pages}{1--14}.
\newblock


\bibitem[\protect\citeauthoryear{Fourneau, Pacanowski, and Barla}{Fourneau et~al\mbox{.}}{2024}]%
        {fourneau2024}
\bibfield{author}{\bibinfo{person}{Gary Fourneau}, \bibinfo{person}{Romain Pacanowski}, {and} \bibinfo{person}{Pascal Barla}.} \bibinfo{year}{2024}\natexlab{}.
\newblock \showarticletitle{Interactive Exploration of Vivid Material Iridescence based on Bragg Mirrors}.
\newblock  \bibinfo{volume}{43}, \bibinfo{number}{2} (\bibinfo{year}{2024}).
\newblock


\bibitem[\protect\citeauthoryear{Gitlina, Guarnera, Dhillon, Hansen, Lattas, Pai, and Ghosh}{Gitlina et~al\mbox{.}}{2020}]%
        {gitlina2020}
\bibfield{author}{\bibinfo{person}{Yuliya Gitlina}, \bibinfo{person}{Giuseppe~Claudio Guarnera}, \bibinfo{person}{Daljit~Singh Dhillon}, \bibinfo{person}{Jan Hansen}, \bibinfo{person}{Alexander Lattas}, \bibinfo{person}{Dinesh Pai}, {and} \bibinfo{person}{Abhijeet Ghosh}.} \bibinfo{year}{2020}\natexlab{}.
\newblock \showarticletitle{Practical measurement and reconstruction of spectral skin reflectance}.
\newblock \bibinfo{journal}{\emph{Computer Graphics Forum}} \bibinfo{volume}{39}, \bibinfo{number}{4} (\bibinfo{year}{2020}), \bibinfo{pages}{75--89}.
\newblock
\urldef\tempurl%
\url{https://doi.org/10.1111/cgf.14055}
\showDOI{\tempurl}


\bibitem[\protect\citeauthoryear{Guarnera, Gitlina, Deschaintre, and Ghosh}{Guarnera et~al\mbox{.}}{2022}]%
        {guarnera2022}
\bibfield{author}{\bibinfo{person}{Giuseppe~Claudio Guarnera}, \bibinfo{person}{Yuliya Gitlina}, \bibinfo{person}{Valentin Deschaintre}, {and} \bibinfo{person}{Abhijeet Ghosh}.} \bibinfo{year}{2022}\natexlab{}.
\newblock \showarticletitle{{Spectral Upsampling Approaches for RGB Illumination}}. In \bibinfo{booktitle}{\emph{Eurographics Symposium on Rendering}}, \bibfield{editor}{\bibinfo{person}{Abhijeet Ghosh} {and} \bibinfo{person}{Li-Yi Wei}} (Eds.). \bibinfo{publisher}{The Eurographics Association}.
\newblock
\showISBNx{978-3-03868-187-8}
\showISSN{1727-3463}
\urldef\tempurl%
\url{https://doi.org/10.2312/sr.20221150}
\showDOI{\tempurl}


\bibitem[\protect\citeauthoryear{Jeon, Baek, Yi, Fu, Dun, Heidrich, and Kim}{Jeon et~al\mbox{.}}{2019}]%
        {minkim}
\bibfield{author}{\bibinfo{person}{Daniel~S. Jeon}, \bibinfo{person}{Seung-Hwan Baek}, \bibinfo{person}{Shinyoung Yi}, \bibinfo{person}{Qiang Fu}, \bibinfo{person}{Xiong Dun}, \bibinfo{person}{Wolfgang Heidrich}, {and} \bibinfo{person}{Min~H. Kim}.} \bibinfo{year}{2019}\natexlab{}.
\newblock \showarticletitle{Compact snapshot hyperspectral imaging with diffracted rotation}.
\newblock \bibinfo{journal}{\emph{ACM Trans. Graph.}} \bibinfo{volume}{38}, \bibinfo{number}{4}, Article \bibinfo{articleno}{117} (\bibinfo{date}{jul} \bibinfo{year}{2019}), \bibinfo{numpages}{13}~pages.
\newblock
\showISSN{0730-0301}
\urldef\tempurl%
\url{https://doi.org/10.1145/3306346.3322946}
\showDOI{\tempurl}


\bibitem[\protect\citeauthoryear{LeGendre, Bladin, Kishore, Ren, Yu, and Debevec}{LeGendre et~al\mbox{.}}{2018}]%
        {legendre2018}
\bibfield{author}{\bibinfo{person}{Chloe LeGendre}, \bibinfo{person}{Kalle Bladin}, \bibinfo{person}{Bipin Kishore}, \bibinfo{person}{Xinglei Ren}, \bibinfo{person}{Xueming Yu}, {and} \bibinfo{person}{Paul Debevec}.} \bibinfo{year}{2018}\natexlab{}.
\newblock \showarticletitle{Efficient multispectral facial capture with monochrome cameras}.
\newblock In \bibinfo{booktitle}{\emph{ACM SIGGRAPH 2018 Posters}}. \bibinfo{publisher}{Association for Computing Machinery}, \bibinfo{address}{New York, NY, USA}.
\newblock


\bibitem[\protect\citeauthoryear{LeGendre, Yu, and Debevec}{LeGendre et~al\mbox{.}}{2016}]%
        {legendre2016}
\bibfield{author}{\bibinfo{person}{Chloe LeGendre}, \bibinfo{person}{Xueming Yu}, {and} \bibinfo{person}{Paul Debevec}.} \bibinfo{year}{2016}\natexlab{}.
\newblock \showarticletitle{Efficient Multispectral Reflectance Function Capture for Image-Based Relighting}. In \bibinfo{booktitle}{\emph{Color and Imaging Conference}}, Vol.~\bibinfo{volume}{24}. Society for Imaging Science and Technology, \bibinfo{pages}{47--58}.
\newblock


\bibitem[\protect\citeauthoryear{Pharr, Jakob, and Humphreys}{Pharr et~al\mbox{.}}{2023}]%
        {pharr2023}
\bibfield{author}{\bibinfo{person}{Matt Pharr}, \bibinfo{person}{Wenzel Jakob}, {and} \bibinfo{person}{Greg Humphreys}.} \bibinfo{year}{2023}\natexlab{}.
\newblock \bibinfo{booktitle}{\emph{Physically based rendering: From theory to implementation}}.
\newblock \bibinfo{publisher}{MIT Press}.
\newblock


\bibitem[\protect\citeauthoryear{Toisoul, Dhillon, and Ghosh}{Toisoul et~al\mbox{.}}{2018}]%
        {toisoul2018acquiring}
\bibfield{author}{\bibinfo{person}{Antoine Toisoul}, \bibinfo{person}{Daljit~Singh Dhillon}, {and} \bibinfo{person}{Abhijeet Ghosh}.} \bibinfo{year}{2018}\natexlab{}.
\newblock \showarticletitle{Acquiring spatially varying appearance of printed holographic surfaces}.
\newblock \bibinfo{journal}{\emph{ACM Transactions on Graphics (TOG)}} \bibinfo{volume}{37}, \bibinfo{number}{6} (\bibinfo{year}{2018}), \bibinfo{pages}{1--16}.
\newblock


\bibitem[\protect\citeauthoryear{Werner, Velinov, Jakob, and Hullin}{Werner et~al\mbox{.}}{2017}]%
        {werner2017scratch}
\bibfield{author}{\bibinfo{person}{Sebastian Werner}, \bibinfo{person}{Zdravko Velinov}, \bibinfo{person}{Wenzel Jakob}, {and} \bibinfo{person}{Matthias~B Hullin}.} \bibinfo{year}{2017}\natexlab{}.
\newblock \showarticletitle{Scratch iridescence: Wave-optical rendering of diffractive surface structure}.
\newblock \bibinfo{journal}{\emph{ACM Transactions on Graphics (TOG)}} \bibinfo{volume}{36}, \bibinfo{number}{6} (\bibinfo{year}{2017}), \bibinfo{pages}{1--14}.
\newblock


\bibitem[\protect\citeauthoryear{Yan, Ha{\v{s}}an, Walter, Marschner, and Ramamoorthi}{Yan et~al\mbox{.}}{2018}]%
        {yan2018rendering}
\bibfield{author}{\bibinfo{person}{Ling-Qi Yan}, \bibinfo{person}{Milo{\v{s}} Ha{\v{s}}an}, \bibinfo{person}{Bruce Walter}, \bibinfo{person}{Steve Marschner}, {and} \bibinfo{person}{Ravi Ramamoorthi}.} \bibinfo{year}{2018}\natexlab{}.
\newblock \showarticletitle{Rendering specular microgeometry with wave optics}.
\newblock \bibinfo{journal}{\emph{ACM Transactions on Graphics (TOG)}} \bibinfo{volume}{37}, \bibinfo{number}{4} (\bibinfo{year}{2018}), \bibinfo{pages}{1--10}.
\newblock


\bibitem[\protect\citeauthoryear{ZEISS}{ZEISS}{2024}]%
        {zeiss}
\bibfield{author}{\bibinfo{person}{ZEISS}.} \bibinfo{year}{2024}\natexlab{}.
\newblock \bibinfo{title}{ZEISS Spectrometer Modules: Compendium of products, electronic components and software solutions}.
\newblock \bibinfo{howpublished}{\url{https://asset-downloads.zeiss.com/catalogs/download/spc/adb036b4-b2e5-48d2-ae76-3903ddfe9c06/BR_Spectrometer_Modules_EN.pdf}}.
\newblock


\end{thebibliography}
\end{document}